\documentclass[conference]{IEEEtran}
\IEEEoverridecommandlockouts
\usepackage{cite}
\usepackage{titlesec}
\usepackage{amsmath,amssymb,amsfonts}
\usepackage{algorithmic}
\usepackage{graphicx}
\usepackage{float}
\usepackage{textcomp}
\usepackage{xcolor}
\usepackage{url}

\titlespacing*{\subsubsection}{0pt}{1.25ex plus 1ex minus .2ex}{0.4em}

\def\BibTeX{{\rm B\kern-.05em{\sc i\kern-.025em b}\kern-.08em
    T\kern-.1667em\lower.7ex\hbox{E}\kern-.125emX}}
\begin{document}

\title{Graph Network Models To Detect Illicit Transactions In Block Chain	}

% \author{
% \IEEEauthorblockN{Hrushyang Adloori}
% \IEEEauthorblockA{\textit{UFID: 3399 1217} \\
% \textit{University of Florida}\\
% adloorih@ufl.edu}
% }
% \author{
% \IEEEauthorblockN{Abhijith Chandra Mergu, Author 2 Name, Author 3 Name, and Author 4 Name}
% \IEEEauthorblockA{\IEEEauthorrefmark{1}Department/Institution 1\
% Email address 1}
% \IEEEauthorblockA{\IEEEauthorrefmark{2}Department/Institution 2\
% Email address 2}
% }
\author{
\IEEEauthorblockN{Hrushyang Adloori}
\IEEEauthorblockA{
\textit{University of Florida}\\
\textit{UFID:  8677 6220}\\
adloorih@ufl.edu}
\and
\IEEEauthorblockN{Vaishnavi Dasanapu}
\IEEEauthorblockA{
\textit{University of Florida}\\
\textit{UFID: 9005 6271}\\
vdasanapu@ufl.edu}
\and
\IEEEauthorblockN{Abhijith Chandra Mergu}
\IEEEauthorblockA{
\textit{University of Florida}\\
\textit{UFID: 3399 1217}\\
a.mergu@ufl.edu}
}

\maketitle

\begin{abstract}
The use of cryptocurrencies has led to an increase in illicit activities such as money laundering, with traditional rule-based approaches becoming less effective in detecting and preventing such activities. In this paper, we propose a novel approach to tackling this problem by applying graph attention networks with residual network-like architecture (GAT-ResNet) to detect illicit transactions related to anti-money laundering/combating the financing of terrorism (AML/CFT) in blockchains. We train various models on the Elliptic Bitcoin Transaction dataset, implementing logistic regression, Random Forest, XGBoost, GCN, GAT, and our proposed GAT-ResNet model. Our results demonstrate that the GAT-ResNet model has a potential to outperform the existing graph network models in terms of accuracy, reliability and scalability. Our research sheds light on the potential of graph related machine learning models to improve efforts to combat financial crime and lays the foundation for further research in this area.
\end{abstract}
\vspace{5pt}

\section{Introduction}

\subsection{Motivation}
\begin{figure}[htbp]
\centering
\includegraphics[width=0.5\textwidth]{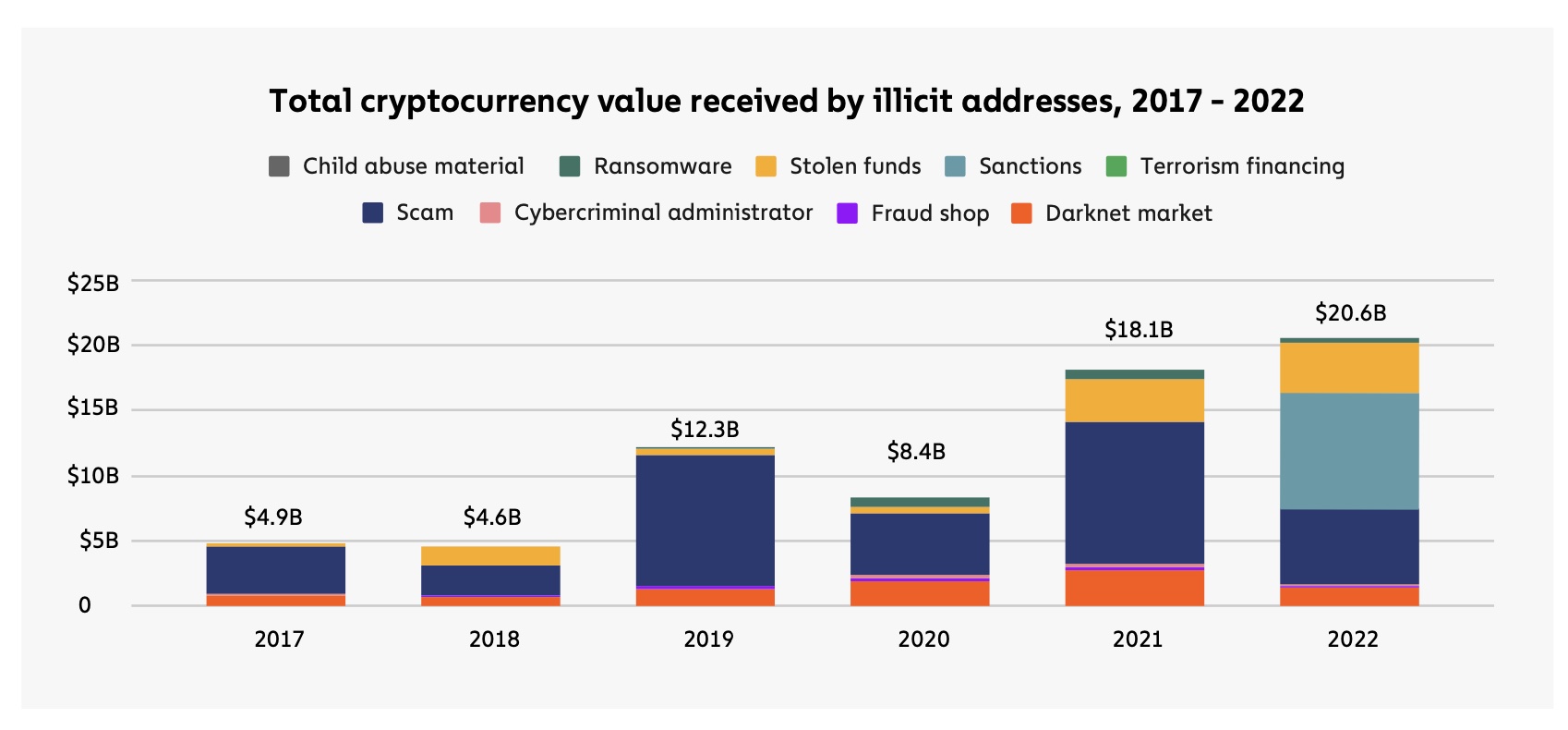}
\caption{Total cryptocurrency value received by illicit addresses}
\label{fig:Chainalysis Data}
\end{figure}

In recent years, blockchain technology has gained significant attention due to its decentralized and immutable nature, which has the potential to revolutionize various industries, including finance, supply chain management, and healthcare. However, as with any innovative technology, it also presents new challenges and risks. In particular, blockchain-based cryptocurrencies have become a preferred tool for conducting illicit activities, such as money laundering, terrorist financing, and drug trafficking. According to the latest report by Chainalysis\cite{b1}, illicit transaction volume in cryptocurrencies rose to an all-time high of \$20.6 billion in 2022, despite the market downturn. Figure \ref{fig:Chainalysis Data} presents the data related to the total cryptocurrency value received by the illicit addresses in the last five years. This highlights the urgent need for effective anti-money laundering (AML) measures to combat such criminal activities.

Existing AML methods primarily rely on traditional financial systems, which may not be applicable to blockchain-based transactions. Thus, there is a need for advanced techniques that can detect and prevent illicit transactions on blockchain networks. In this context, machine learning (ML) has shown promising results in analyzing large amounts of blockchain data and identifying patterns of illicit activities. In particular, graph-related models such as Graph Convolutional Networks (GCN), Graph Attention Networks (GAT), and GAT ResNet have shown great potential in analyzing the complex network structure of blockchain transactions.

This paper aims to investigate the effectiveness of different ML models, including GCN, GAT, and GAT ResNet, in detecting illicit transactions on the blockchain network. We utilize the publicly available Elliptic data set, which contains labeled transactions related to illicit activities. Our approach involves feature engineering on the transaction graph, followed by training and testing the ML models on this data set. The results of our experiments will provide valuable insights into the effectiveness of graph-related ML models for detecting illicit transactions on the blockchain network.

In summary, this paper addresses the critical issue of detecting illicit transactions on blockchain networks using advanced ML techniques. Our research has practical implications for developing effective AML measures that can counteract the growing trend of cryptocurrency-related criminal activities.

\subsection{Basics of Money Laundering in Blockchain}
The process of money laundering boils down to disguising the rewards from illegal activities as legitimate funds. The process involves turning in the illegitimate funds and making it appear as if it is earned from legitimate means. 

On a broader perspective it can be classified into three stages\cite{b2}: Placement, Layering and Integration. The initial placement phase involves placing the illegally earned wealth into dumpy financial business of which the books can be easily manipulated, often invested in smaller amounts to avoid attracting unnecessary attention. The later layering phase basically involves minimizing the association of the funds from the original owner in the most efficient and undetectable ways possible. The final integration phase is the reward reaping phase where all illegitimate funds are finally available to use in the legitimate economy.

In the context of blockchain transactions, the following are the common techniques\cite{b3} used by money launderers:

\textbf{1. Tumblers:} Tumblers are online services that mix legitimate and illegitimate cryptocurrency transactions to create a virtual fund that can be sent to a new address, effectively disrupting the transactional link between wallets.

\textbf{2. The OTC Market:} Over-the-counter (OTC) trading allows users to transact cryptocurrencies without an organized exchange, making it an attractive option for money launderers due to its anonymity and lack of regulatory oversight.

\textbf{3. Privacy Coins:} Privacy coins such as Monero, Dash, and Zcash were created to provide anonymity to cryptocurrency transactions, but their alleged secrecy is still questionable, with a significant number of trades being traceable.

\textbf{4. Decentralized Exchanges:} Decentralized exchanges (DEXs) allow users to control their private keys and transact cryptocurrencies without intermediaries or counterparty risks, making them difficult to regulate and potentially appealing to money launderers.

\textbf{5. Retail Purchases Using Cryptocurrencies:} Criminal proceeds can be laundered by purchasing high-value assets such as real estate, cars, or jewelry using cryptocurrency and reselling them on the open market.

\textbf{6. Mining as a Front:} Criminals can funnel illicit funds to a cash-intensive business like cryptocurrency mining, mixing illegal coins with regular mining proceeds to hide their income and complete the laundering cycle.

\section{Existing Systems}

This section discusses various systems and techniques in use currently to detect and associate illicit transactions to aid Anti Money Laundering in Block chain.

\subsection{KYC and Identity Verification}

Anti-Money Laundering (AML) and Know Your Customer (KYC) Regulations - Before a consumer is permitted to use a financial service, their identity is identified and verified through the KYC process. AML laws are intended to stop criminals from passing off stolen money as lawful revenue. Financial institutions are required
by these requirements to keep an eye on and alert regulatory authorities to any questionable transactions. To stop and identify money laundering, exchanges and other service providers in the cryptocurrency industry must adhere to KYC and AML requirements. Each client is given a risk score based on their behavior and other pertinent characteristics when risk scoring is used. This may include elements like the origin nation, previous transaction history, and other details.

Financial institutions may more easily identify high-risk clients and keep a closer eye on their behavior by giving them risk scores \cite{b11}. Anti-Money Laundering (AML) laws and procedures are a collection of rules intended to thwart the use of financial systems for money laundering operations. 

Making unlawful gains seem as though they were obtained via legal ways to hide their actual source is known as money laundering. The integrity and security of financial institutions throughout the world are threatened by this activity, which is frequently linked to criminal gangs like narcotics traffickers and terrorists \cite{b12}. To identify suspicious transactions, software systems used in the detection of financial fraud frequently use rule engines that rely on predefined rules created by subject- matter experts. A series of rules is applied by the rule engine one at a time, and a transaction is only flagged as fraudulent if one of the rules is triggered. Such technologies' primary flaw is their incapacity to identify unidentified fraud patterns. 

Experts must first discover and specify the underlying techniques before adding new rules to the rule base to recognize novel circumstances. Algorithms for machine learning and data mining allow for the possibility of identifying new behavioral patterns and spotting questionable activity in financial transactions. The fundamental issue is the uneven nature of the data, where the proportion of fraudulent transactions to genuine ones is rather low. Effective fraud detection models are difficult to create because labeled datasets for anomalous cases are either nonexistent or difficult to collect.

\subsection{Conventional rule-based approaches}

These systems, as their name implies, rely on hard- coded rules that are configured to flag transactions when they satisfy specific requirements. Such regulations may be created by following industry best practices, such as blocking transactions from a single account from occurring repeatedly within a short period of time, transactions coming from VPNs, or transactions coming from dangerous locations, analyzing fraudulent transactions that have been caught or prevented, and developing new rules to cover all of their suspicious characteristics.

Almost all imperative programming languages have "if-else" expressions, which are often used to represent the rules and are simple to understand. They mimic how a person would handle a transaction by checking to see whether it fits any of the dangerous patterns specified in the rules, and if it does, blocking it or sending it for manual human review. Stakeholders trust them since they approach the assignment in the same manner that they would do it themselves, which is one of the reasons why their presence is still quite strong.

Conventional rule-based solutions for money laundering detection and prevention in cryptocurrency transactions have limits since they can't keep up with the way that money laundering methods are continually developing \cite{b13}. The dynamic and ever-evolving nature of cryptocurrencies and money laundering necessitates a more
complex and adaptable strategy, even though rule-based solutions might be useful in some circumstances.

\subsection{Machine Learning Based Approaches}

As the rise of machine learning models grew to be more and more accurate in recent years, especially dealing with this particular problem with availability of a structured dataset. Many approaches have been pursued to include machine learning models that can be trained to detect such illicit transactions. Phetsouvanh et al. \cite{b6} proposed a deep learning approach for classifying blockchain transactions as either legitimate or illicit. The authors used a combination of convolutional and recurrent neural networks to extract features from the transaction data.

A study conducted by Zhang et al \cite{b5}. proposed a hybrid approach for detecting money laundering activities on the Bitcoin blockchain. The study initiated the use of graph networks dealing with this problem. But,  the study does not provide a detailed analysis of the false positive and false negative rates of the approach, which may limit its practical utility in real-world settings. 

Collaboration of Weber et al. and the Elliptic Co.\cite{b4} produced a well-structured research on introducing the use of graph convolutional networks and its variants like EvolveGCN. This study provided a strong foundation as the actual inception of the Graph Networks and rooted for them due their higher level of association with the model of the data. However, the study could produce Graph Networks that could match the performance of RandomForest or neither be close to it. 

As part of research work, we experiment on various kinds of machine learning models implementing it on the Elliptic dataset. Also, proposing a novel approach to implement a variant of Graph Attention Network that can perform better than existing Graph Network approaches. As the variant, we chose GAT combined with Residual Network-like architecture that has better correlation with the pertained problem. As far as our research goes, there is no study which attempted to implement Residual Network variation of GAT on classifying the Illicit/Licit transactions in the domain of blockchain.

\section{Graph Networks}
Graph Networks are a type of machine learning model that has proven to be a powerful tool for modeling complex relationships between entities in a dataset.\cite{b8}

\subsection{Advantages of Graph Networks}
Graph Networks provide various advantages over classic machine learning models such as logistic regression and decision trees when it comes to the problem of detecting illegal transactions in an elliptic dataset using machine learning techniques. Graph Networks are able to capture the complex interdependencies between the elements in the dataset in a form that is clearly interpretable because they represent the entities in the dataset as nodes in a graph and the relationships between them as edges in the graph. [8] This capability to model complex relationships is particularly useful in the detection of illicit transactions, where the relationships between entities may be intricate and difficult to represent using traditional machine learning models. This ability to model complex relationships is particularly useful in the detection of illicit transactions.

In addition, Graph Networks can tolerate incomplete or missing data, which is a typical problem in many applications that are used in the real world.

Graph Networks use a method called graph convolution to propagate information through the graph and fill in missing data. This allows them to handle incomplete data more effectively than typical machine learning models, which rely on the assumption that all data is complete. In addition, Graph Networks are able to manage data on a vast scale as well as data that is diverse, which makes them suited for the extensive and complicated elliptic dataset. Graph Networks are able to scale to enormous datasets thanks to the distributed representation of the graph that they utilize. Additionally, Graph Networks can manage a wide variety of data formats.

Because of these benefits, Graph Networks are especially relevant when it comes to the context of identifying illegal transactions. As a consequence of this, Graph Networks are positioned to become increasingly essential in a wide variety of applications that are used in the real world. Graph Convolutional Networks (GCN), Graph Attention Network (GAT), and a variation of GAT that is of prime importance in this research and is the subject of this paper's primary attention, GAT infused Residual Network model (GAT-ResNet) are the three primary types of Graph Network models that will be discussed in this particular research study. In the following sections, we will devote considerable attention to a discussion of each of these graph networks in detail.

\subsection{Graph Convolutional Networks (GCN)}

Graph Convolutional Networks, often known as GCNs, are a special kind of neural network that can handle data that is arranged as a graph. The concept of employing convolutional operations to extract features from the nodes and edges of a graph is the foundation of the GCN architecture, which is built on this concept. \cite{b9} Traditional convolutional networks are designed to function on regular grids, whereas GCN are designed to function on irregular graphs. Because of this, GCN are ideally suited for jobs that include graph-structured data.

GCN can be used to represent the data as a graph and learn characteristics from it in the context of detecting illegal transactions in the elliptic dataset. This can be accomplished through the usage of the elliptic dataset. The data will first be displayed in the form of a graph in the first stage of this process. Each entity in the dataset, such as a Bitcoin address, can be represented as a node in the graph, and the relationships between entities, such as Bitcoin transactions between addresses, can be represented as edges in the graph.

The construction of the graph is the next step to take. Defining a set of rules to connect nodes and edges based on the features of the dataset is one way to accomplish this goal. These rules can be derived from the properties of the dataset. For instance, if two addresses have jointly taken part in a transaction, then there is a possibility that there will be a connection between them in the form of an edge. Following the construction of the graph, the next step is to define features for each node and edge in the structure. These attributes can be generated from a variety of sources, such as the transaction history of a specific location or the characteristics of the parties that are participating in a transaction. Another such source is the history of a transaction at a particular address.

One of the benefits of using GCN to detect illegal transactions is that it is able to capture the intricate links that exist between the many entities that make up the dataset. Traditional machine learning models may have difficulty capturing the subtleties of these relationships, but GCN is able to do so because it learns from the interactions that occur between nodes and edges as information is propagated around the graph. In addition, GCN is resilient in the face of missing or partial data, which qualifies them for the task of dealing with the noisy and incomplete data that are present in the elliptic dataset.

Another one of GCN's many strengths is its capacity to manage data on a massive scale and in a variety of formats. The elliptic dataset is a great example of this type of data because it has a wide variety of different types of entities and also has a large scale. GCN is able to do this by utilizing a distributed representation of the graph, which enables it to take into account the properties that are specific to the various kinds of entities.

In conclusion, Graph Convolutional Networks, also known as GCN, are an effective tool for processing graph-structured data, which is present in a variety of applications that are used in the real world. GCN may be used to represent the data as a graph and learn features from it, which makes them well-suited for managing the complex, heterogeneous, and noisy data that is present in the elliptic dataset. One use case for this is the detection of unlawful transactions in the elliptic dataset.

\subsection{Graph Attention Network (GAT)}
Graph Attention Networks, also known as GATs, are a specific variety of artificial neural network that are intended for use in the analysis of data that has been organized in the form of a graph. They are able to collect data not just at the level of individual nodes but also at the level of the graph as a whole in its whole.\cite{b10} This is accomplished by utilizing an attention mechanism that gives weights to nodes in the graph that are located in close proximity to one another.

GAT can be particularly beneficial when applied to the challenge of discovering illicit transactions in the elliptic dataset because it can detect intricate linkages between entities that may be seeking to conceal their activities. This is because GAT can identify entities that are attempting to conceal their activities. Because it is able to learn from both local and global patterns in the graph, GAT is particularly well-suited to the task at hand.

GAT employs a self-attention mechanism, in contrast to the conventional attention mechanisms, which grants it the ability to assign varying weights to the various neighbors of a given node. Because of this, the network is able to learn more intricate and significant relationships between the items that make up the graph.

In order to apply GAT to the elliptic dataset, the data must first be structured in the form of a graph, with each entity being represented as a node in the graph and the interactions between entities being represented as edges. It is possible to extract the attributes that are used to represent nodes and edges from a variety of sources, such as the history of transactions and characteristics of the parties that are participating in a transaction.

One of GAT's features is its capacity to manage incomplete and noisy data, which is particularly relevant for detecting unlawful transactions. Additionally, GAT is able to handle a wide variety of data types. In addition, GAT is able to analyze data on a wide scale as well as data from a variety of sources, which makes it a strong candidate for the elliptic dataset.

In conclusion, GAT is an effective neural network design that has the capability of capturing the intricate interactions that exist between the elements that make up the elliptic dataset. Because of its capability to do local as well as global pattern analysis, deal with missing or partial data, and manage large-scale and diverse data sets, it is an efficient tool for discovering illegal transactions.

\subsection{GAT Residual Network (GAT-ResNet)}
Graph Attention Networks, also known as GAT, are capable of successfully extracting information from graphs at both the node-level and the global-level. The GAT-ResNet architecture is a variant of GAT that combines the power of GAT with residual connections modeled after ResNet in order to increase the performance of the model. This form of GAT is one of the variations of GAT. 

The utilization of many GAT layers, each with residual links to the next, is required for the GAT-ResNet protocol. The model is able to learn the residual mapping between layers thanks to the residual connections, which helps relieve the vanishing gradient problem that is typically seen in deep learning models.

The GAT-ResNet architecture can be used to the graph representation of the dataset in order to find fraudulent transactions in the elliptic dataset. Each entity is represented as a node, and the relationships that exist between them are shown as edges in the diagram. There are a variety of characteristics that can be used to represent the nodes and edges of the graph. These characteristics include the number of transactions, the value of transactions, the categories of transactions, and the number of incoming and outgoing transactions for each entity.

Because of its many strengths, the GAT-ResNet architecture is well suited for the task of uncovering illegal dealings in the elliptic dataset. It is able to capture the intricate connections that exist between the entities in the dataset, as well as the local and global patterns that exist in the graph. It is not affected by data that are absent or incomplete, which is a significant benefit taking into account the magnitude and variety of the elliptic dataset. In addition, GAT-ResNet is capable of managing large-scale and diverse data, a capability that is essential for identifying illegal transactions in real-world scenarios.

In conclusion, the GAT-ResNet architecture is an effective method for uncovering illegal dealings inside the elliptic dataset. Its ability to extract node-level attributes and global-level information from the graph, together with its tolerance to missing data and its ability to manage large-scale and heterogeneous data, make it a viable solution for tackling this difficult challenge.

\section{Methodology}

The goal of the study is to suggest a novel system for detecting suspicious transactions and patterns of behavior associated with money laundering operations in bitcoin transactions using machine learning algorithms. The study involved a thorough literature research of the existing systems in the area, and the results are presented in the sections above. The shortcomings of the existing work is discussed in the later sections by comparing the proposed novel model's performance against them.
\subsection{Process Outline}

\begin{figure}[htbp]
\centering
\includegraphics[width=0.5\textwidth]{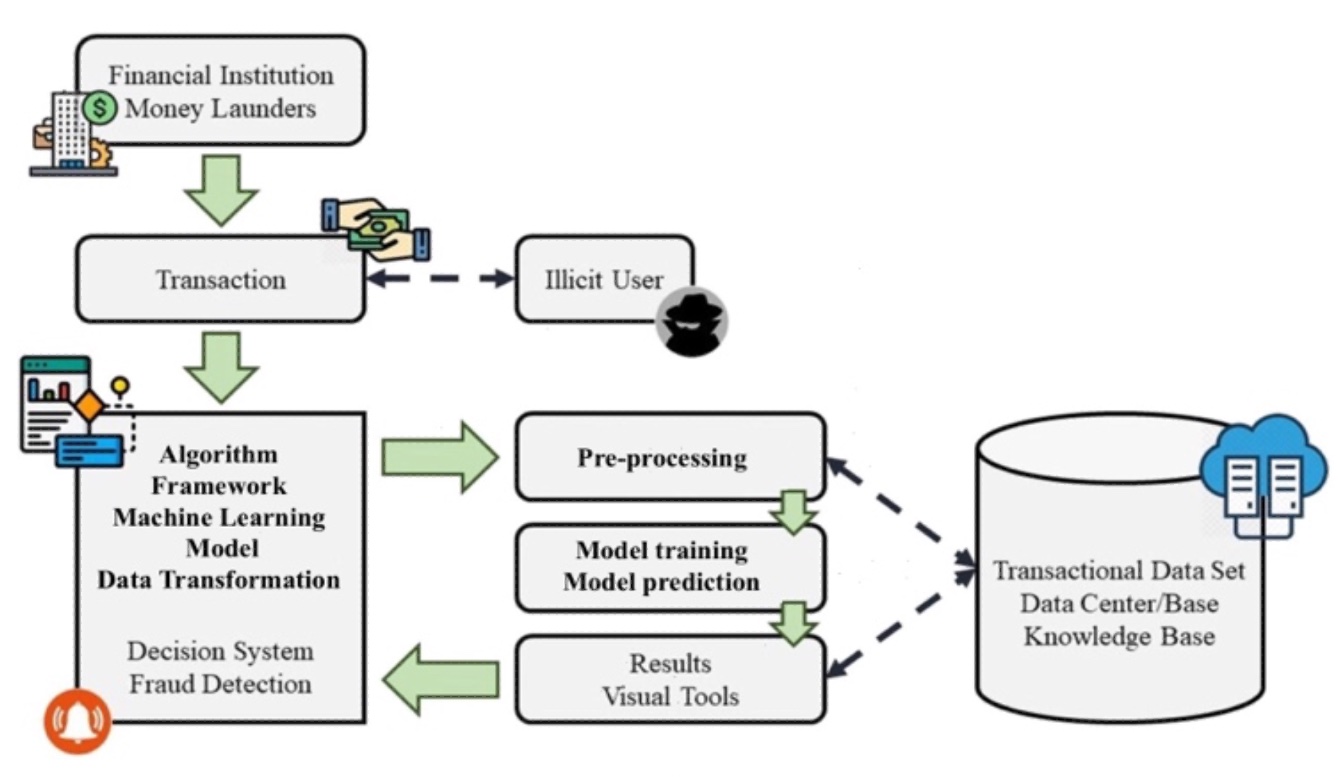}
\caption{Process Outline}
\label{fig:ProcessOutline}
\end{figure}

The Figure \ref{fig:ProcessOutline} explains overall flow of the framework. The entities participating in the exchange of crypto initiate transactions, of which few of them could be illicit. This transaction data is fed into a novel system based on machine learning techniques which would help in classifying the illicit transactions thus helping in preventing the transactions related to money laundering or other illegal means.

\subsection{Elliptic Data}
We use the blockchain dataset provided by Elliptic \cite{b7}, a provider of technology-based solutions for risk analysis, compliance, and blockchain analytics. Elliptic focuses in offering financial institutions, cryptocurrency companies, and governmental organizations compliance and forensics solutions for cryptocurrencies.

The elliptic data set labels/classifies bitcoin transactions into licit and illicit transactions. The transactions falling under licit include those originated from exchanges, wallet providers, miners, licit services, etc. And the illicit transactions are the ones which are originated from scams, malware, terrorist organizations, ransomware, Ponzi schemes, etc.

The dataset contains of transaction graphs. A node in the graph in represents a transaction and an edge represents flow of bitcoins from one transaction to other. There are a total of 49 graphs obtained from bitcoin blockchain at different timestamps.

\begin{figure}[htbp]
\centering
\includegraphics[width=0.5\textwidth]{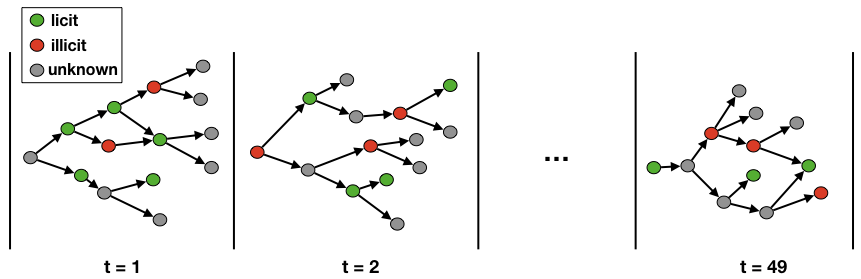}
\caption{Different connected components of elliptic dataset based on time step}
\label{fig:EllipticData}
\end{figure}

The graph is made of 203,769 nodes and 234,355 edges. Two percent (4,545) of the nodes are labelled class1 (illicit). Twenty-one percent (42,019) are labelled class2 (licit). The remaining transactions are unknown. Each graphs represents two weeks of data about the transactions in that time period. There are 166 features associated with each node, of which 94 features represent local information and other 72 represent aggregated information, obtained using various transactions information.

\subsection{Splitting the Dataset}
In this study, we aimed to evaluate the effectiveness of our model on the Elliptic Data Set, which involves detecting illicit transactions over time. To ensure a fair evaluation of our model's performance, we divided the data into two sets: training and test. We used a 70:30 ratio, where 70\% of the data was used for training and the remaining 30\% for testing.

To capture the temporal nature of the data, we employed a temporal split where the first 34 time steps were used for training and the last 15 for testing. This split reflects the nature of the task since it involves detecting illicit transactions over time. Our approach allowed us to evaluate our model's ability to generalize to new and unseen data.

The use of a temporal split is critical to ensure that our model is not overfitting to the training data, which could result in poor performance on unseen data. By reserving a portion of the dataset for testing, we were able to evaluate the model's ability to adapt to new and unseen data over time. Moreover, using a temporal split allowed us to train our model in an inductive setting, where the model is trained on a portion of the data and tested on a different set of data. This approach is essential for real-world scenarios where the data is continually changing.

\subsection{Training Traditional Models}

To evaluate the performance of various classification models on the task of licit/illicit prediction, we experimented with three widely used approaches. The first one was Logistic Regression, which we implemented using the default parameters provided by the scikit-learn Python package. Next, we used Random Forest, which is also available in scikit-learn. For this model, we set the number of estimators and the maximum number of features to 50. Finally, we tested XGBoost, which is another popular classifier available in scikit-learn. We set the number of estimators to 50, maximum depth to 100, and the random state to 15.

\subsection{Building and Training Graph Models}

But our main interest still remains in experimenting and implementing the Graph Networks. As these graph networks have relevance and much more scope in being a possible solution in tackling the problem when run in real-time. This belief is already established by stating how graph networks have greater scope to capture the transactions and their relations. Since these models have higher reliability and have greater scope dealing with this problem, we have experimented with different graph network models and their variants to tackle the classification problem. \\

\subsubsection{Training Graph Convolutional Networks (GCN) }

During the training process, we utilized a 2-layer Graph Convolutional Network (GCN) model for 1000 epochs, employing the Adam optimization algorithm with a learning rate set at 0.001. After fine-tuning the hyperparameters, we established the optimal size for the node embeddings to be 100.

The objective of this experiment was binary classification, with the dataset containing two imbalanced classes. In the context of Anti-Money Laundering (AML), the minority class, which represents illicit transactions, is of greater significance. To address this imbalance and prioritize the minority class, we employed a weighted cross-entropy loss function during the training of the GCN model. This approach allowed us to assign higher importance to the illicit samples.

Following the hyperparameter tuning process, we determined the most suitable ratio for the licit and illicit classes to be 0.3/0.7. This decision was made to ensure that the model would be better equipped to handle the class imbalance and accurately classify the data, with a particular focus on the detection of illicit transactions.\\

\subsubsection{Training Graph Attention Networks (GAT) }

We used a 2-layer Graph Attention Network (GAT) model with 8 heads and set the size of the node embeddings to 100 after hyperparameter tuning to classify a set of data into two categories. The task was binary classification, with the dataset containing two imbalanced classes. To address this imbalance and prioritize the minority class, we employed a weighted cross-entropy loss function during the training of the GAT model. We trained the GAT model for 1000 epochs using the Adam optimizer with a learning rate of 0.001 and a patience of 50 to prevent overfitting and improve the generalization of the model. After fine-tuning the hyperparameters, we determined that the optimal ratio for the licit and illicit classes was 0.3/0.7.\\

\subsubsection{Training GAT-ResNet Model}

Our goal was to optimize the GAT model to accurately classify the data while accounting for the class imbalance, using a weighted cross-entropy loss function and fine-tuning the hyperparameters to achieve the best possible results. We utilized a GAT-ResNet model, which combines the strengths of Graph Attention Network (GAT) and Residual Network (ResNet) architectures, to classify a set of data into two categories. This model is particularly advantageous for the problem at hand, as it has been shown to outperform traditional GAT models in terms of accuracy and robustness.

After hyperparameter tuning, we set the size of the node embeddings to 100 (hidden channels) and used a 3-layer GAT-ResNet network with 4 heads. The layers have residual connections introducing the ResNet-like architecture into GAT. The task was binary classification, with the dataset containing two imbalanced classes. To address this imbalance and prioritize the minority class, we employed a weighted cross-entropy loss function during the training of the GAT-ResNet model.

We trained the GAT-ResNet model for 1000 epochs using the Adam optimizer with a learning rate of 0.001 and a patience of 50 to prevent overfitting and improve the generalization of the model. After fine-tuning the hyperparameters, we determined that the optimal ratio for the licit and illicit classes was 0.3/0.7.

To optimize the GAT-ResNet model to accurately classify the data while accounting for the class imbalance, we used a weighted cross-entropy loss function and fine-tuning the hyperparameters to achieve the best possible results.

\begin{figure}[htbp]
\centering
\includegraphics[width=0.4\textwidth]{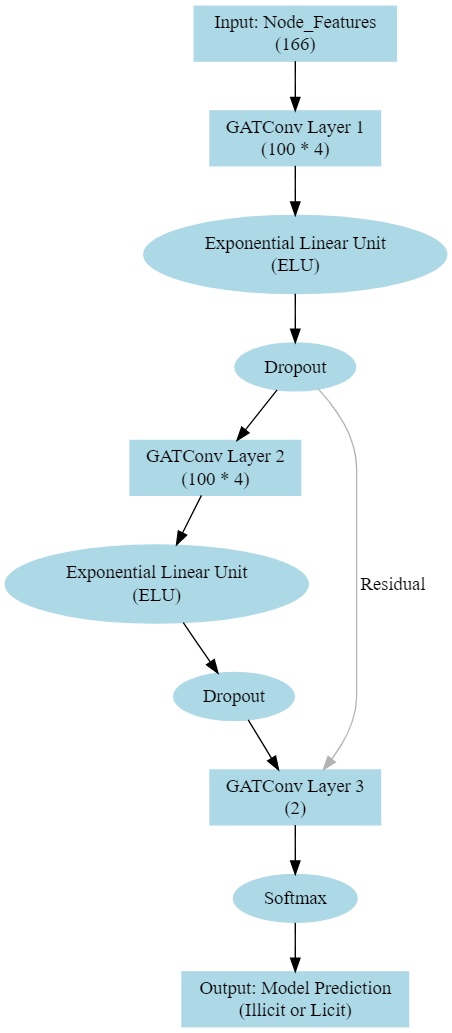}
\caption{GAT-ResNet Architecture}
\label{fig:GATResNetArchitecture}
\end{figure}

\subsection{GAT-ResNet Architecture}

\begin{figure*}[htbp]
\centering
\includegraphics[width=0.7\textwidth]{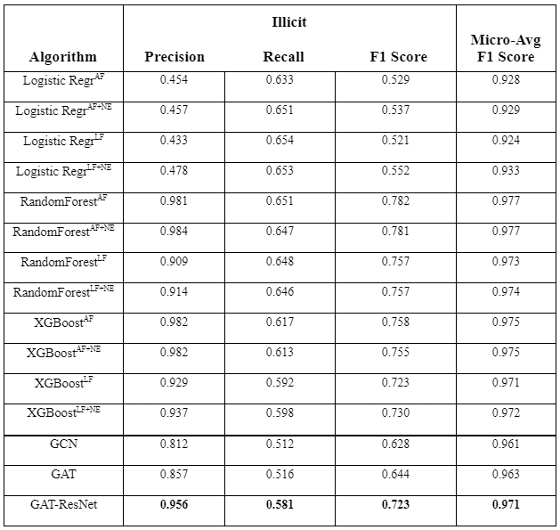}
\caption{Illicit Transaction Prediction}
\label{fig:IllicitTrans_Pred}
\end{figure*}

The GATResNet model is a custom PyTorch module designed to handle the classification of elliptic Bitcoin transaction datasets. This model combines the strengths of Graph Attention Network (GAT) and Residual Network (ResNet) architectures, providing better accuracy and robustness in the classification of imbalanced datasets.

Figure \ref{fig:GATResNetArchitecture} shows the detailed architecture of the GAT-ResNet model used. The model consists of three GAT layers, with the first two layers mapping the input node features to hidden channels using attention heads, and the final layer mapping the output of the second layer to two output classes. The model also includes an ELU activation function and dropout with a probability of 0.5 after the first and second layers, as well as a residual connection between the outputs of the first and second GAT layers.

If the use\_skip flag is set to True, a skip connection is added between the input node features and the output of the final GAT layer, using a weight matrix initialized with Xavier normal initialization. Finally, a softmax activation function is applied to the output to obtain class probabilities.

The embed method of the GATResNet class returns the output of the first GAT layer, which can be used as node embeddings for further analysis or visualization. Overall, the GATResNet model is a powerful tool for accurately classifying elliptic Bitcoin transaction datasets, with the added benefit of being able to handle imbalanced datasets with greater accuracy and robustness.

\section{Results and evaluation}

In the case of the traditional machine learning models, the super scripts indicate the feature considerations while they are being evaluated. ‘AF’ and ‘LF’ denote All Features and Local Features respectively. All Features includes all the 166 features associated with each node or transaction while the Local Features include the selected 94 that represent the more locally associated features of that particular node. ‘NE’ indicates node embedding representing a low-dimensional vector representation of each node (i.e., entity or address) in the Bitcoin transaction network. These node embeddings capture the characteristics of each node based on its local and global network structure, as well as its attributes, such as transaction volume, degree, and clustering coefficient.  \\

There are several metrics used in the study to evaluate and compare the performances of various machine learning models discussed. The following metrics are used for the initial evaluation:

\subsubsection{Precision (P)} The proportion of actual positive results relative to the total number of projected positive results is what precision measures. The capacity of the classifier to correctly identify as positive just those samples that actually are positive is one intuitive definition of precision. For instance, a classifier that simply labels everything as positive would have a precision of 0.5 in a test set that was evenly balanced between positive and negative examples (50 percent each). A test would be considered to have a precision of 1 if it did not produce any false positives, or in other words, if it identified only the genuine positive results. In a nutshell, the precision of a classifier is directly proportional to the number of false positives it produces.

\[ P = \frac{TP}{TP + FP}\]

\begin{center}
TP - True Positives and FP - False Positives
\end{center}

\subsubsection{Recall (R)} Recall counts the percentage of real positives among all positives. The number of test samples that were actually labeled as positive can be thought of as recall. A classifier with a recall of 1.0 but a lower precision would return positive for every sample, regardless of whether it is actually positive. A classifier's recall increases with the number of false negatives it produces.

\[ R = \frac{TP}{TP + FN}\]
\begin{center}
TP - True Positives and FN - False Negatives
\end{center}

\subsubsection{F1 Score} Harmonic mean of precision and recall for a more balanced summarization of model performance. F1 score balances importance of precision and recall.

\[ F1 =  2 \times \frac{P \times R}{P + R}\]

\begin{center}
P - Precision and R - Recall
\end{center}

\subsubsection{Micro-Avg F1 Score} By adding the sums of the True Positives (TP), False Negatives (FN), and False Positives (FP), micro averaging determines the global average F1 score. In order to determine our micro F1 score, we first add the corresponding TP, FP, and FN values across all classes and plug them into the F1 formula to obtain the micro average F1.
\\

The evaluations are initially based on the commonly used evaluation metrics mentioned above - Precision, Recall, F1 score and micro average F1. From results of table shown in Figure \ref{fig:IllicitTrans_Pred}, we can observe RandomForest and XGBoost models clearly outperform all the other models. Although the accuracies of these have no significant differences, the RandomForest model has substantial advantage of having better recall and F1 values than compared to the XGBoost model. 

The GAT-ResNet model, which was proposed as an efficient solution to detect illicit transactions was compared with the above discussed two graph network models, the GCN and GAT. The performance of these models was evaluated initially using the same metrics mentioned above.

\begin{figure}[h]
  \centering
  \fbox{\includegraphics[width=0.45\textwidth]{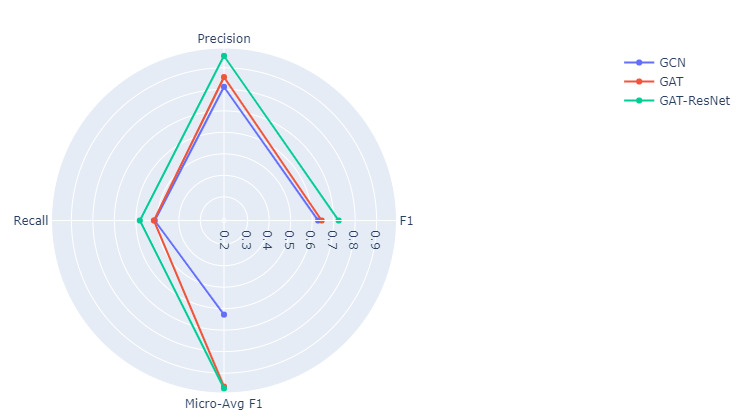}}
  \caption{Various metric comparison of graph network models}
  \label{fig:spiderGraph}
\end{figure}

The results from Table in Figure \ref{fig:spiderGraph} of the study clearly indicate a significant improvement in all the metrics using the GAT-ResNet model when compared to the existing graph network models. The scores of this model were found to be near the range of that of RandomForest and XGBoost models, which are known to be highly effective for classification problems especially the blockchain transactions as claimed by \cite{b4}.

To further evaluate the performance of GAT-ResNet against other graph network models, another metric known as Matthews correlation coefficient (MCC) was used. The metric is calculated using this equation.

\[MCC = \frac{(TP \times TN - FP \times FN)}{\sqrt{(TP+FP)(TP+FN)(TN+FP)(TN+FN)}}\]

While detecting the anomaly remains the primary focus of these models, avoiding the false flags of a genuine transaction is equally significant. So, considering accuracies alone could be a misleading and imperfect method to evaluate these models. Hence evaluating based on MCC score would be appropriate in this scenario. Matthews correlation coefficient (MCC) is a metric commonly used in binary classification problems to evaluate the performance of a model. MCC takes into account true positive, true negative, false positive, and false negative predictions and returns a value between -1 and 1, where 1 indicates perfect predictions, 0 represents random predictions, and -1 indicates total disagreement between the predictions and the true labels.

\begin{figure}[htbp]
\centering
\includegraphics[width=0.5\textwidth]{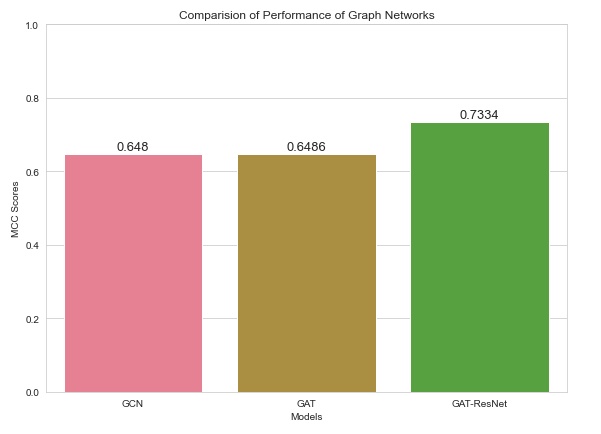}
\caption{MCC value comparison of graph network models}
\label{fig:resultGraph}
\end{figure}

MCC takes into account both true positives and true negatives, which is particularly relevant in imbalanced datasets as is the case with the used Elliptic dataset. A high MCC score indicates that the model has correctly predicted both the illicit and licit transactions, while a low MCC score indicates that the model has made more errors in its predictions. As shown in the Figure \ref{fig:resultGraph} GAT-ResNet achieves a higher MCC score than GCN and GAT, further establishing the reliability of the model.

\section{Future Scope}
The study resulted in improving the performance of graph networks but couldn't outperform the powerful and proven models like RandomForest and XGBoost. However, with emerging research on how effectively graph networks can be engineered, with appropriate technical resources and engineering reduces the gap between these models. Underlying mechanics and flags that trigger predictions of these models is still a black box, advancing the graph networks in this domain could potentially hold the possibilities of this question. 

Also, ensemble modelling of traditional models with these graph networks could be the perfect blend to tackle this problem. Further, strenghtening AML in vulnerable networks like blockchain. In future, deploying such models combined with better alternative like Self-Sovereign Identity (SSI) instead of KYC can used. This maintains the certain level of anonymity while keeping AML methods actionable and trackable.
\section{Conclusion}

Graph Networks have high grounds to cover in order to compete with RandomForest. But graph networks having closer connection on how the problem is posed, having more scope to be technically engineered with combinations and variants of such models could easily replace the prominent RandomForest model. The  results from our novel approach of implementing Graph Attention Network infused with Residual Network like architecture proves the basis of this claim.

\vspace{12pt}
\end{document}